\documentclass[12pt,letterpaper]{article}

\usepackage[margin=1in]{geometry}
\usepackage{setspace}
\usepackage{times}

\usepackage{xcolor}
\usepackage{hyperref}
\usepackage{graphicx}
\usepackage{subcaption}
\usepackage{longtable}

\usepackage{apacite}

\begin{document}

{\singlespacing
\title{\vspace{-0.5cm}\Large\textbf{Hybrid Panels: Toward Human–AI Collaboration\\ in Survey Research}}
\author{\normalsize\textbf{Julia Romberg\textsuperscript{1}, Tobias Gummer\textsuperscript{1,2}, Gabriella Lapesa\textsuperscript{1,3}, Tanja Kunz\textsuperscript{1},}\\ \normalsize\textbf{and Claudia Wagner\textsuperscript{1,4}}\\\\
\normalsize\textsuperscript{1}GESIS - Leibniz Institute for the Social Sciences\\
\normalsize\textsuperscript{2}Heidelberg University\\  
\normalsize\textsuperscript{3}Heinrich Heine University of Düsseldorf\\
\normalsize\textsuperscript{4}RWTH Aachen University\\
}
\date{\normalsize Contact: \{firstname.lastname\}@gesis.org}
\maketitle}

\vspace{-1cm}
\begin{abstract}
\noindent
Large-scale population surveys are essential for generating robust social and scientific insights, yet they face significant challenges, including declining response rates, increasing data collection costs, long delays between data collection and data provision, and the risk of nonresponse bias. 
Advances in artificial intelligence (AI) have opened up new opportunities for AI-supported survey infrastructures where the goal is to overcome these challenges without limiting the data quality.
A promising AI-enabled survey infrastructure for which we build a first pilot is a hybrid panel. A hybrid panel is a longitudinal AI-enabled survey which allows to iteratively improve the alignment between large language models (LLMs) and the population they aim to simulate and use the errors to inform the design and implementation of the next survey wave (e.g., inform the participant recruitment, assignment of questions to participants). It incorporates both human participants and LLMs as fundamental elements of its design.

In this research note, we introduce the concept of a hybrid panel by providing a definition and outlining an overarching framework, spanning data collection to data validation.~We~detail results from a first pilot study to illustrate (open) challenges that we identify for hybrid panels.
\end{abstract}

\section{Introduction}

Population surveys are a longstanding and imperative means to measure opinions, attitudes and values, with application in various fields, from measuring attitudes, values, and behavior for social science research over public opinion polling and to market research for product and service development.
Despite their importance, survey research faces several persistent challenges, including declining response rates \shortcite{brick2013explaining, williams2018trends, de2018international,luiten2020survey}, increasing data collection costs \shortcite{olson2021transitions, wolf2021conducting}, long delays between data collection and data provision, and the risk of nonresponse bias, particularly among important population subgroups and hard-to-reach populations \shortcite{schanze2023response, stein2026education}.
Recent advancements in artificial intelligence (AI) and, in particular, large language models (LLMs)  have sparked optimism regarding their potential to address challenges in survey research by promoting efficiency, accessibility, and flexibility \shortcite{rothschild2024opportunities}.
The use cases in which AI can support survey research are manifold and span the entire survey lifecycle \shortcite{rothschild2025}, ranging from questionnaire design and survey item generation \shortcite{gotz2024let} to AI-assisted interviewing \shortcite{wuttke-etal-2025-ai, 10.1145/3381804} and the automated coding of open-ended responses \shortcite{Heyde_2025}.

Yet, realizing the potential of AI in survey research also requires considering the infrastructures within which these applications would be embedded.
A fundamental requirement for conducting traditional population surveys are data collection infrastructures, which provide the organizational and methodological backbone for collecting high-quality data and making them available for scientific use~(e.g., to enable analyses of the general population over time). 
Over the past decades, such infrastructures have been established for major longitudinal national and international survey programs, including repeated cross-national studies (e.g., the US General Social Survey, the European Social Survey, the World Values Study) and panel studies (e.g., the Panel Study of Income Dynamics, the German Socio-Economic Panel, and Understanding Society). Beyond enabling the systematic collection of data over time, these infrastructures provide researchers with access to already recruited, probability-based samples and established methodological standards, protocols, and quality control procedures. 

Commercial providers such as \href{https://www.qualtrics.com/}{Qualtrics}, \href{https://xpolls.ai/}{Xpolls}, and \href{https://personapanels.com/}{PersonaPanels} have already begun to build up structures to offer synthetic panels generated from AI models. These solutions---although promising faster and less costly data collection---face severe challenges related to transparency, accountability, and ethical compliance. These concerns are especially salient in social science research because open science, reproducibility, and the ability to independently evaluate data quality are essential requirements for scientific integrity. What is more, recent research has also called into question the overall readiness of the technology for full synthesis \shortcite{NEURIPS2024_515c6280, tjuatja-etal-2024-llms, von2025vox}.

In this research note, we introduce the concept of a \textbf{hybrid panel for social science research} by providing a definition and outlining an overarching framework \textbf{to facilitate the systematic development and testing of AI-enabled survey infrastructures}. We define a hybrid panel as a longitudinal survey infrastructure in which responses are provided by both human participants and AI models, for example through planned missingness designs where missing data is (in part) imputed via LLMs,  the augmentation of human responses with AI-generated input, or human validation of AI-generated responses.
The longitudinal design reflects the need for ongoing maintenance and human participation in developing suitable AI methods for imputation, ideally with continuous validation of AI-generated responses against data collected from the same human participants over time.
The hybrid panel bridges the gap between traditional panels based on human samples and fully synthetic panels, acknowledging both the current deficiencies of synthetic approaches such as the mismatch with human behavior \shortcite{NEURIPS2024_515c6280, tjuatja-etal-2024-llms, von2025vox} and the promise of AI for survey research \shortcite{Argyle_Busby_Fulda_Gubler_Rytting_Wingate_2023}. We hereby extend on previous research that integrates responses from human participants and AI within theoretical frameworks, including prediction-powered inference for valid statistical analysis when human observations are supplemented by AI-generated responses \shortcite{science_Angelopoulos_2023,krsteski2025valid}, mixed subject designs combining observations from both sources \shortcite{doi:10.1177/00491241251326865}, and adaptive resource allocation via informed sampling methods that are frequently used in the area of active machine learning to reduce model errors \shortcite{settles2009active}.
To illustrate our framework, we introduce our ongoing research endeavor of building up a hybrid panel for social science research. We describe our design of the hybrid panel for social science research that we started piloting, discuss the methodological and practical challenges that we encountered and present initial findings of our pilot study which covers the human participant recruitment step.

\textit{Before turning to the remainder of the paper, note that the authors of this research note come from different disciplinary backgrounds~(Survey Research, Computer Science, Computational Social Science, Computational Linguistics) and have non-overlapping positions on the very important debate on AI in survey research. What we share, however, is the firm conviction that AI cannot fully replace humans as the object of study of survey research, and at the same time that the extent of the applicability of a technology to a research domain is a matter of evaluation and experimentation and it requires an ambitious and rigorous research agenda with a long-term time plan. This is precisely the contribution that we want to make with the hybrid panel outlined in this research note, which can only be developed further through the feedback we hope to receive from a community that is, at the time of writing, extremely split and for good reasons.}

%%%
\section{Implementing a Hybrid Panel for Social Science Research}

\subsection{Definition}

Survey responses obtained exclusively from human participants and those generated exclusively by AI systems (the so-called in-silico panels) represent the two ends of a continuum. Our proposed hybrid panel falls within this continuum by integrating both human participants and LLMs into a common longitudinal survey infrastructure in which they collaborate, e.g., by supplementing or validating each other’s input.
More specifically, we define a hybrid panel as a group of human participants and AI models that repeatedly contribute to surveys or annotation tasks over time. 
Rather than treating AI as a replacement for human participants, the hybrid panel enables various forms of interplay between humans and AI. 
For example, human responses or human annotations can be used to train, calibrate, and validate AI models that then may support users in filling out a survey or completing an annotation task (e.g., by suggesting entries for items that users skip). Another option would be to test planned-missingness designs (e.g., split-questionnaire designs) where the questionnaire is divided into subsets of questions, and different respondents receive different subsets. The planned missing items can be answered by AI models and the assignment can be rotated across subsequent waves. Over time, respondents may eventually answer all modules, which allows one to validate AI-generated answers while each individual wave remains shorter.

Such hybrid designs may allow novel adaptive survey designs, re-distribution of data collection budget, e.g., to invest into the recruitment of hard-to-reach populations, and advancing the state-of-the-art in both LLM alignment and its intersection to survey research.
The survey research community may also benefit from practical evaluations on a more fundamental level, as it will allow to sharpen and adapt crucial notions such as the distinction between imputation and simulation, and an update of the notion of data quality.  

It is to be noted that our definition deliberately extends beyond traditional survey tasks to include data annotation tasks, the process of labeling data snippets according to predefined concepts: providing LLMs with access not only to opinions, attitudes or values on a more conceptual level but also with concrete evaluations of data snippets that provide textual manifestations of these high-level concepts.
For example, a survey item on attitudes toward populism can be complemented with several short texts that respondents evaluate as populist or non-populist. This design moves beyond abstract self-reports by capturing how individuals apply the concept in concrete contexts. The resulting data provide a richer and more behaviorally grounded representation of respondents' perspectives, which is particularly well suited for LLM-based modeling due to its reliance on semantic judgments and language-based reasoning. From the survey participant perspective, integrating annotation in traditional surveys may also end up to be more engaging than ticking scales or producing open-ended answers. Annotation within our hybrid survey infrastructure can be implemented as a form of gamification which may increase the appeal of surveys for socio-demographic groups who perceive traditional survey formats as unengaging. 

%%%
\subsection{Overall Framework}

\begin{figure}[ht]
    \centering
    \includegraphics[width=1\linewidth]{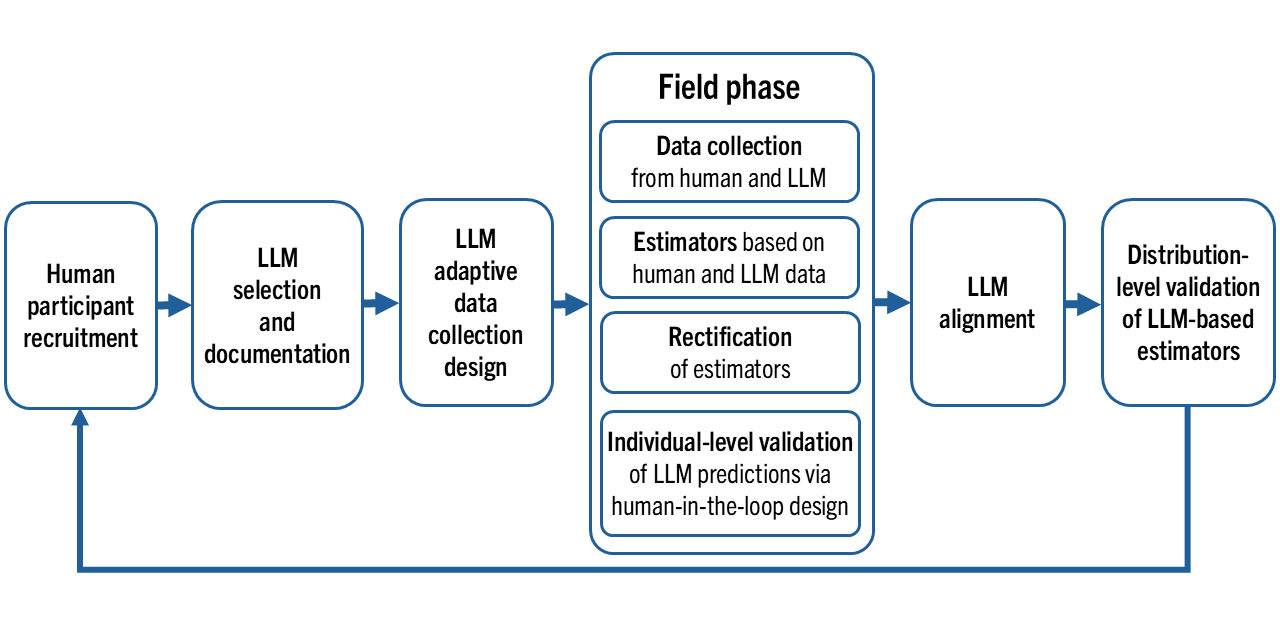}
    \caption{Phases of the hybrid panel.}
    \label{fig:framework}
\end{figure}

Figure \ref{fig:framework} illustrates the different phases in the implementation of our hybrid panel.
The first phase concerns the \textbf{recruitment of human participants} where the challenges summarized in the representation arm of the Total Survey Error Framework apply \shortcite{groves2004survey}. Theories on survey participation decisions \shortcite{dillman2014internet,groves2000leverage} of potential respondents suggest that cost-benefit considerations play a key role in these decisions, taking into account the general context and the design of a survey. In the case of hybrid panels, we argue that it is important to consider the conditions under which participants are willing to collaborate with and share their data for AI model training. Factors such as public interest, trust, data privacy, security, and transparency are expected to positively affect their willingness to participate \shortcite{Waind_2020}. The impact of further design options---such as the degree of autonomy of AI, the sensitivity of data shared by participants for AI learning, the permissible sensitivity of data that the AI outputs, and the amount of financial compensation---remains uncertain in the novel context of hybrid panels and is investigated in our pilot study (see Section \ref{sec:pilot}).
Although ethical considerations related to the potential replacement of human participants by LLMs and the resulting reduction or loss of compensation are relevant, we do not advocate for it nor do we anticipate this transition occurring in the near future.

The second phase concerns the \textbf{LLM(s) selection and documentation}.
To ensure transparency, accountability, and ethical compliance, preference should be given to open and transparent models whose training data, model architecture, and weights are publicly available for inspection. We will use existing documentation procedures such as the recently released LLM-checklist \shortcite{Feuerriegel2026LLMChecklist} to document the LLM selection and usage processes. Furthermore, the selected LLMs should be calibrated for the specific survey task. This is an open challenge, since so far no specialized model exists for survey research and general-purpose LLMs reveal many forms of undesirable response biases and inconsistencies. For example, LLM responses are sensitive to changes in the question wording and show social desirability biases \shortcite{tjuatja-etal-2024-llms}. To what extent these biases can be reduced by fine-tuning LLMs on a large amount of survey data is an open empirical question (see, e.g., \shortciteA{ceron2026politicalcontentllmspre}).

The third phase focuses on designing and implementing what we call \textbf{LLM adaptive data collection design}. The goal is to determine which information should be imputed and which should be collected from which respondents to maximize estimator accuracy keeping the data collection costs within reasonable limits.\footnote{Although the LLM adaptive survey design aims to optimize resource allocation, its goal is not to remove human input, but to reposition it where it adds the most value.}
In traditional panels, adaptive survey design varies survey procedures between groups of respondents during and after data collection based on information gathered about respondents, response patterns, costs, or sample composition \shortcite{schouten2017adaptive}. For hybrid panels, methods developed for machine learning settings (e.g., uncertainty sampling, expected error reduction) \shortcite{settles2009active} ---where large amounts of unlabeled data coexist with a smaller amount of labeled ground-truth data---may provide a promising complement.

During the \textbf{field phase}, human and LLM data are collected. Based on the LLM adaptive data collection design, we decide for which humans we collect empirical data and for whom data is simulated.  Empirical and synthetic data are used to create survey estimators for the variables. Statistical methods such as design-based supervised learning \shortcite{Naoki2023}, prediction-powered inference \shortcite{science_Angelopoulos_2023} or confidence-driven inference \shortcite{gligoric-etal-2025-unconfident} will be used to rectify the biased LLM estimators. Human-in-the-loop designs can be used to assess the validity of individual-level predictions, as evaluation by those whose perspectives are modeled offers the most effective and ethically grounded way to assess AI simulations. 
In practice, an LLM responds to selected survey items or question groups, and these responses (or a strategically sampled subset of responses) are presented to human participants for assessment rather than requiring them to complete the items themselves. This allows researchers to identify systematic discrepancies between human and LLM responses, quantify prediction uncertainty, and iteratively improve the alignment of the LLM models. At the same time, it gives participants an active role in evaluating how their views are represented, increasing transparency, accountability, and trust in AI-assisted survey methodologies.

After the field phase, the selected \textbf{LLMs will be aligned} to the data that was collected during this iteration. Fine-tuning (i.e., further training a pre-trained model on a specific dataset and task) or in-context learning (i.e., supplying the model with examples or instructions directly in the input) are possible options.
As discussed before, the longitudinal nature is a key characteristic of a hybrid panel. Having survey waves over time enables us to validate the LLM component of the hybrid panel in between waves. In the last phase, we assess the \textbf{validity of LLM predictions at the distribution level} by comparing them against human-informed survey estimates based on data that were unavailable during model training and evaluation. By relying on unpublished survey data\footnote{GESIS is involved in several panel initiatives in which a considerable amount of time typically elapses between data collection and publication.}, we minimize the risk of data leakage\footnote{Data leakage occurs when information that should not be available to the model accidentally influences its development (e.g., the survey to be used as test dataset). This can make the model appear more accurate than it truly is, resulting in misleading performance estimates.} and ensure that the resulting performance estimates reflect genuine predictive capability.
We will use the validation step to provide detailed performance reports of the LLM component of the panel (e.g., which models and which parameter settings worked well, for which subgroups do we get the lowest performance, which questions cannot be well predicted) and infer design decisions about the next panel waves from these reports.

%%%
\subsection{Pilot Study}
\label{sec:pilot}

We conducted a pilot study with the crowdworking platform Prolific\footnote{\href{https://www.prolific.com/}{https://www.prolific.com/}} as a widely adopted and cost-efficient environment.
In the first phase of the pilot, we address the step of human participant recruitment and validating the sample itself, before, in later phases, we will begin implementing the subsequent steps of the hybrid panel.
The invitation to participate to the survey was sent to 6,032 eligible participants (German residents who are fluent in German).
Data collection took place from June 1 to June 7, 2026. To ensure a diverse sample, we collected 300 respectively 301\footnote{We opened 300 places for each of the four time points. A submission error led to the collection of 301 responses in one of the batches.} responses at different time points: twice during weekdays (morning and evening), on a public holiday (noon), and on the weekend (morning).
Participants were incentivized for participation with a German minimum wage (13.90 EUR per hour) and gave informed consent. 
A total of 1,260 participants started the survey, of whom 1,201 completed it, 45 exited at an early stage and 14 were filtered out by the platform due to exceeding the maximum time allowed (35 minutes) without completing the task. The average completion time among respondents with valid responses was 11 minutes and 33 seconds. 

Our main research questions are:
\begin{itemize}
    \item \textbf{RQ1.} Do respondents already use AI in their work on Prolific?
    \item \textbf{RQ2.} Would respondents be willing to participate in a hybrid panel?
    \item \textbf{RQ3.} Which respondents would take part in a hybrid panel?
\end{itemize}

RQ1 investigates whether respondents use ChatGPT and inter alia to answer survey items. Understanding such behavior is crucial for a controlled hybrid panel, otherwise it remains shallow what true human-written and what AI-generated responses are.
To assess this non-optimal answering behavior, we implemented a self-reported survey question on whether respondents use AI support in their work on Prolific\footnote{While we use the term ``AI'' to simplify understanding for a general audience, the listed answer options refer to applications of LLMs.}. Overall, we found that only 6\% of participants reported using AI for response generation or multiple-choice selection. Most respondents, 73\%, reported no AI usage, while the remainder reported to use AI for minor language revisions.
An additional spot check of the writing style in free-text response supports the low numbers of self-reported AI-use for our study. A recent Prolific study provides additional evidence of low AI prevalence on crowdworking platforms \shortcite{gordon_rothschild_affonso_sulik_hauser_pepin_jones_2026}.
These findings suggest that our sample can be used to implement the further steps of our proposed hybrid panel pilot.

\begin{figure}[ht!]
    \centering
    \begin{subfigure}[b]{0.45\textwidth}
        \centering
        \includegraphics[width=\textwidth]{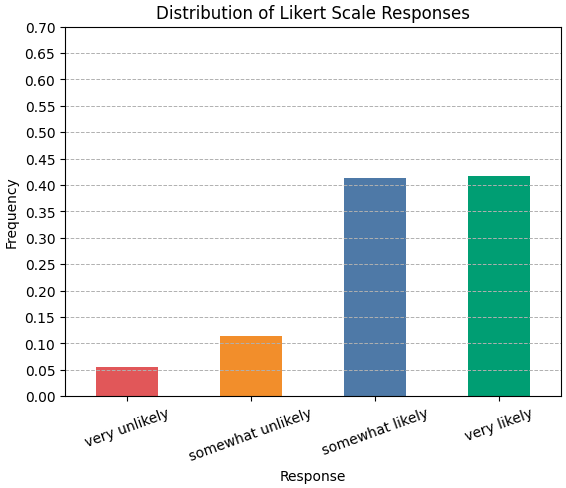}
        \caption{Participation in a human panel.}
        \label{fig:likert_a}
    \end{subfigure}
    \begin{subfigure}[b]{0.45\textwidth}
        \centering
        \includegraphics[width=\textwidth]{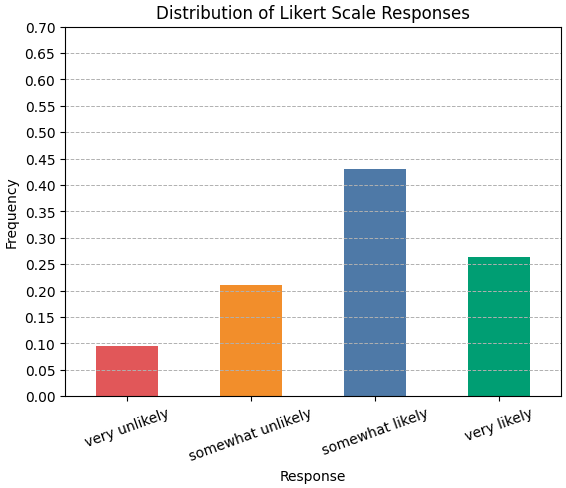}
        \caption{Participation in a hybrid panel.}
        \label{fig:likert_b}
    \end{subfigure}
    \caption{Distribution of responses regarding the likelihood of participating in a regular human panel and a hybrid panel, as asked sequentially in the survey.}
    \label{fig:panel-vs-hybrid}
\end{figure}

RQ2 assesses the respondents' general willingness to participate in a traditional or a hybrid panel (see Figure \ref{fig:panel-vs-hybrid}): Whereas 83\% of the respondents reported to likely participate in a traditional panel (somewhat likely: 41\%, very likely: 42\%), 69\% reported that they likely would participate in a hybrid panel (somewhat likely: 43\%, very likely: 26\%). Although these results are positive overall, they indicate that participation rates are lower for a hybrid panel than for a traditional panel and that design decisions are needed to improve them.
\begin{figure}[ht!]
    \centering
    \begin{subfigure}[b]{0.45\textwidth}
        \centering
        \includegraphics[width=\textwidth]{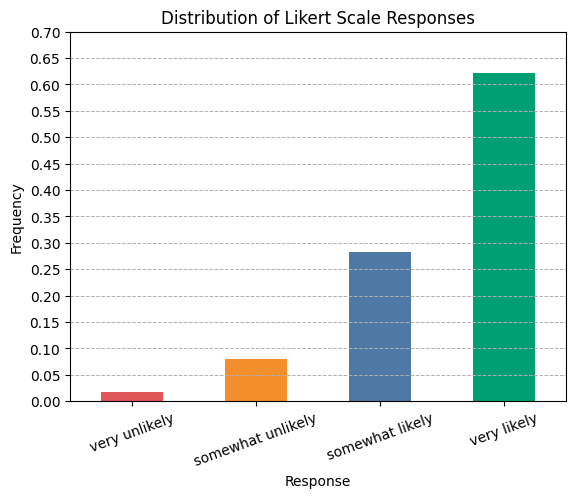}
        \caption{Low data sensitivity, low AI autonomy, and high financial compensation.}
        \label{fig:likert_c}
    \end{subfigure}
    \begin{subfigure}[b]{0.45\textwidth}
        \centering
        \includegraphics[width=\textwidth]{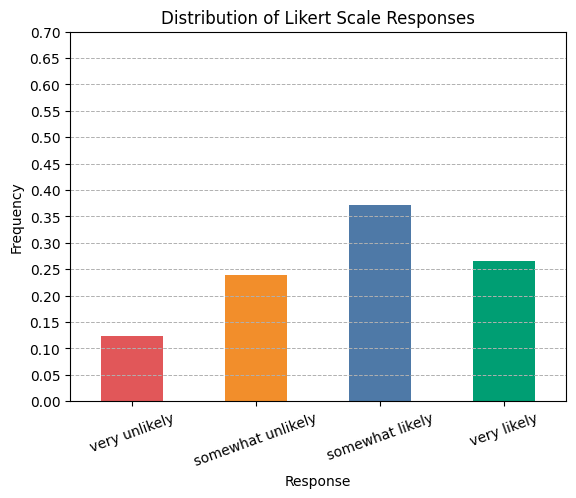}
        \caption{High data sensitivity, high AI autonomy, and low financial compensation.}
        \label{fig:likert_d}
    \end{subfigure}
    \caption{Distribution of responses regarding the likelihood of participating in two opposing vignette design options for the hybrid panel. Option (a) describes the most conservative scenario, option (b) the most progressive case.}
    \label{fig:vignette}
\end{figure}

To better understand how specific design choices affect participation in the hybrid panel, we included a vignette experiment that varied the conditions of data sensitivity (low vs. high), AI autonomy (low vs. high), and financial compensation (low vs. high). Each participant was presented with two randomly assigned vignettes. Figure \ref{fig:vignette} illustrates the results of the two extremes. The combination of low data sensitivity, low AI autonomy, and high compensation was associated with a substantial increase in the likelihood of stated participation, while the opposite scenario of high data sensitivity, high AI autonomy, and low compensation was perceived more negatively, though a notable share of respondents still reported their willingness to participate. These findings indicate that supporting respondent information need to be developed to convey a high degree of transparency about AI-support in the hybrid panel.

RQ3 then explores potential nonresponse biases when recruiting a hybrid panel. 
In order to approximate nonresponse bias, we compared those who reported that they likely would participate in a hybrid panel to those who reported to be unlikely with respect to their attitudes toward AI (sum factor of ATTARI-12 \shortcite{stein2024attitudes}), political leaning (11-point left-right scale), and socio-demographic variables (gender (0=male, 1=female), education (0=low, 1=high)).
We found that respondents who were likely to participate in a hybrid panel had a more positive attitude toward AI (M=3.63) than those who were unlikely to participate (M=3.13, t=10.15, p$<$0.001). We further found differences in their political leaning, with likely respondents leaning more towards the right (M=4.86 vs. M=4.26, t=4.4, p$<$0.001). With respect to gender, we found a higher share of males among the likely respondents (Chi$^2$=26.55, p$<$0.001). Concerning their education, our data show no differences between likely and unlikely respondents (Chi$^2$=1.37, p$>$0.05).
These findings suggest that nonresponse bias may be a concern when recruiting human participants for a hybrid panel.

%%%
\section{Discussion}

Exploring the potential of AI for survey research is only at its beginning, and it needs a long-term research agenda that takes into account the different components: the human subjects with their preferences and rights  (which we focus on in the pilot study), the AI models and their continuous evaluation (whose features we only briefly touch upon in this research note), and the research community on whose reaction we call with this research note (because the standards for data quality and for robust generalization can only come from a community effort where diverse perspectives and disciplines are represented).
Leveraging AI models for survey research raises expectations of cost reduction and increased speed of data collection.
More efficient data collection enables researchers to respond more swiftly to societal events and to adapt surveys in real time.
Nonetheless, accurately capturing genuine human opinions remains particularly difficult, and the current body of research demonstrates insufficient quality of replicating self-reported human behavior fully synthetically (e.g., \shortciteA{NEURIPS2024_515c6280, tjuatja-etal-2024-llms, von2025vox, pmlr-v202-santurkar23a, durmus2024measuringrepresentationsubjectiveglobal,bulte-terryn-26}).
Discrepancies from the ground truth are particularly consequential in applications where the survey outcomes inform the public or policy makers.
These limitations suggest that the benefits of AI may be best realized by selectively combining human responses with LLM estimators.

Hybrid panels offer a practical framework for such selective integration. They enable selective delegation: where an LLM can reliably simulate human responses for a particular subpopulation or type of question content, a greater share of responses can be delegated to the model, while more complex content or less predictable cases remain in the hands of the humans about whom researchers aim to make generalizations.
Our vision for a hybrid panel allows for in-process validation of AI performance through direct human-in-the-loop involvement. The simulation of opinions, attitudes, or values intrinsic to human beings---phenomena for which ground truth is typically unavailable---is best evaluated by those whose perspectives are being modeled. This design strengthens validation, improves simulation quality through iterative feedback, and promotes transparency and accountability, thereby increasing trust in AI-assisted survey methodologies.
By incorporating humans directly into the process, we can also allow them to have more control over what data the AI is permitted to learn from and what it is allowed to impute. Our framework enables a systematic perspective on the integration of human and AI in surveys tasks and offers a foundational concept upon which future AI-supported survey infrastructures can be created, classified, and evaluated. 

The effort required from respondents will decrease as LLMs become more aligned with human respondents. Lower respondent burden might improve participation rates, measurement error, and retention of respondents in panel settings.  Validating a proposed response usually requires less effort than creating original responses from scratch, especially for complex survey items, such as free-text responses. Initially, human validation should encompass all AI-generated responses to ensure accuracy and model reliability. 
Hard-to-reach populations and systematic nonresponse when recruiting human participants for hybrid panels will remain an open challenge: obtaining at least few data and participants from underrepresented groups is indispensable for model alignment and validation, and we hope that hybrid panels will allow survey effort and budgets to be at least partially invested for this. If this is successful, hybrid panels can augment available human responses of small or hard-to-reach subgroups.  

%%%
\section*{Ethics Statement}
It is not our belief that the measurement of opinions, attitudes, and values inherent to human individuals can or should be fully automated by computational models. Significant concerns arise when synthesized results only are used to answer research questions about society or guide external stakeholders, such as policymakers or the general public, due not only to the severe limitations of current models affecting accuracy (such as hallucinated answers and uncontrolled or hidden bias) but also to the unclear effects on society (such as diminished public acceptance and declining trust in research). 
Our approach presents a solution to overcoming the persistent challenges in survey research (namely declining response rates, increasing data collection costs, and the risk of nonresponse bias) by leveraging the recent advancements of AI while preserving the underlying subject of observation in the process: the human individual.\footnote{Synthesized survey responses may still be useful in laboratory pre-test scenarios, in which limitations can be counted in.}
The ongoing development of the hybrid panel is planned to be monitored by an ethics board and, broadly speaking, is to be seen as a community-driven effort for which this research note and the development of the proposed hybrid panel for Germany is the trigger and also an exchange platform through calls for shared-task-like evaluation (on the AI development side) and empirical reflection on what the desiderata of the survey community are (at which point would survey researchers consider the output of an LLM robust enough to draw generalizations on?). 

An important aspect to conclude this ethics section is the consent of human participants. We are aware that at present, despite the clear definitions in the Prolific survey, it may not be 100\% clear to the participants that what they are consenting to is the fact that AI may partially take over their answer (therefore reducing their compensation). The extent of this understanding is to be further tested for participants who are in principle willing to take part in the hybrid panel. There are however additional panel design considerations that will make this case very unlikely. Should a participant belong to a group for whom the LLMs produce accurate enough predictions for a specific attribute, this participant will still be recruited again at least in the following cases: a) for the smaller validation of these attributes; b) for other attributes and constructs; c) for purely annotation use cases that will still be supported by the infrastructure. It will be a recruitment consideration to make sure that allocation of requests remains fair, and that ``more predictable'' subjects are not penalized as data workers.

\bibliography{custom_bib}

\section{Appendix A}
\subsection{AAPOR Disclosure Standards}

\begin{itemize}
    \item \textbf{First data source:} The survey data was collected by the authors.
    \item \textbf{Data Collection Strategy:} The survey data was collected with an online survey with participant recruitment on the provider Prolific.
    \item \textbf{Research Sponsor and Conductor:} The survey was designed and conducted by the authors and sponsored by their institution. This information will be specified in the non-blinded version of the manuscript.
    \item \textbf{Measurement Tools/Instruments:} Our study uses the nine questions from the questionnaire that are described in detail with instructions and response options in Appendix \ref{app:questions}.
    \item \textbf{Population Under Study:} Participants had to be German residents and fluent in German. They were 18+, based on the properties of Prolific.
    \item \textbf{Methods Used to Generate and Recruit the Sample:} The sample selected is selected with a non-probability method. All participants on Prolific from the population under study were allowed to take part until the quota of 1200 participants was filled. Note that our final sample includes 1201 completed responses due to a technical issue. To diversify the sample, the study was opened at different times and days. Participants were contacted through Prolific. Participation was incentivised.
    \item \textbf{Method(s) and Mode(s) of Data Collection:} Web survey, in German language.
    \item \textbf{Dates of Data Collection:} The data was collected from June 1 through June 6 of 2026.
    \item \textbf{Whether and How the Data Were Weighted:} The data was not weighted.
    \item \textbf{How the Data Were Processed and Procedures to Ensure Data Quality:} All respondent had a 95\% acceptance rate for microtasks on Prolific. Response times were monitored. We analyzed free-text answers to detect bot usage and added a question on AI usage in the questionnaire. Participants could only complete the survey once.
    \item \textbf{IF A PANEL WAS USED: Panel Description}: not applicable
    \item \textbf{IF INTERVIEWERS OR CODERS WERE USED: Interviewer Details}: not applicable 
    \item \textbf{IF ELIGIBILITY SCREENING WAS DONE: Screening criteria and process}: No screening was conducted.
    \item \textbf{Study Stimuli:} not applicable
    \item \textbf{Dispositions or Response or Participation Rates:} The invitation to participate to the survey was sent to 6,032 eligible participants, of which 1260 started the survey. 
1,201 participants completed the survey, while 45 exited at an early stage and 14 were filtered out by the platform due to exceeding the maximum time allowed and non-completion of task.
    \item \textbf{Sample Sizes:} Our sample size is 1201.
    \item \textbf{Measurement and Model Specification:} The analysis was conducted in Python with the pandas package for conducting the t-test and scipy for conducting the chi$^2$-test.
    \item \textbf{A General Statement Acknowledging Limitations of the Design and Data Collection:} Our data collection is conducted with a non-probability method. Respondents recruited through Prolific are likely to be biased towards young, left-leaning and highly educated.
\end{itemize}

\section{Appendix B}

\subsection{Survey Questions}
\label{app:questions}

\subsubsection{AI usage I}

\paragraph{Instructions:}
Jetzt geht es um Ihre Arbeit für Prolific im Allgemeinen.\\

\noindent (Framing 1) Nutzen Sie bei Ihrer Arbeit für Prolific üblicherweise KI-Unterstützung?\\
(Framing 2) Die Nutzung von KI-Tools ist bei Online-Arbeit inzwischen weit verbreitet und kann sehr unterschiedlich aussehen. Haben Sie in Ihrer Arbeit für Prolific schon einmal KI-Unterstützung verwendet?\\
(Framing 3) Menschen nutzen bei Online-Studien unterschiedliche Hilfsmittel, darunter teilweise auch KI-Unterstützung. Haben Sie in Ihrer Arbeit für Prolific schon einmal KI-Unterstützung verwendet?\\
(Framing 4) Für die Qualität unserer Forschung ist eine möglichst genaue Beschreibung Ihrer tatsächlichen Arbeitsweise wichtig. Es gibt dabei keine richtigen oder falschen Antworten. Haben Sie in Ihrer Arbeit für Prolific schon einmal KI-Unterstützung verwendet?\\

\noindent Bitte wählen Sie alles Zutreffende aus.

\noindent Bitte antworten Sie ehrlich. Ihre Antwort hat keinerlei Auswirkungen auf Ihre Teilnahme oder mögliche Vergütung in dieser Studie und wird nicht an Prolific weitergegeben. 

\paragraph{Response options:} Nein, ich nutze üblicherweise keine KI. - Ja, ich nutze KI üblicherweise, um Texte zu verbessern (z. B. Rechtschreibung, Stil oder sprachliche Überarbeitung). - Ja, ich nutze KI üblicherweise, um komplette Antworten generieren zu lassen (z. B. automatische Erstellung von Antworttexten). - Ja, ich nutze KI üblicherweise, um passende Antworten auszuwählen (z. B. Vorschläge einer KI genutzt, um eine der vorgegebenen Antworten auszusuchen). - Ja, anderes:

\subsubsection{AI usage II}

\paragraph{Instruction:} Aus welchen Gründen nutzen Sie üblicherweise KI-Unterstützung bei Ihrer Arbeit für Prolific?

\noindent Ihre ehrliche Antwort hat keinerlei Auswirkungen auf Ihre Teilnahme oder mögliche Vergütung in dieser Studie. Es erfolgt keine Rückmeldung an Prolific.

\paragraph{Response options:} open-ended question

\subsubsection{Panel consent}

\paragraph{Instructions:} Im Rahmen wissenschaftlicher Studien werden oft die gleichen Personen mehrfach befragt – man spricht dann von einem ``Panel''. 

\noindent Wie wahrscheinlich wäre es, dass Sie an einem solchen Panel teilnehmen? 

\paragraph{Response options:} sehr unwahrscheinlich - eher unwahrscheinlich - eher wahrscheinlich - sehr wahrscheinlich

\subsubsection{Hybrid panel consent}

\paragraph{Instructions:} Panels können durch Künstliche Intelligenz (KI) unterstützt werden, z.B. indem KI beim Antworten hilft, Informationen ergänzt oder an Ihrer Stelle antwortet. Solche Formen werden als ``hybride Panels'' bezeichnet. 

\noindent Wie wahrscheinlich wäre es, dass Sie an einem solchen hybriden Panel teilnehmen? 

\paragraph{Response options:} sehr unwahrscheinlich - eher unwahrscheinlich - eher wahrscheinlich - sehr wahrscheinlich

\subsubsection{Vignettes on hybrid panel consent}

\paragraph{Instructions:} Nun stellen wir Ihnen verschiedene Formen eines hybriden Panels vor.  

\noindent Bitte lesen Sie diese Beschreibung sorgfältig und geben Sie anschließend Ihre persönliche Einschätzung ab.\\ 

\noindent In einem hybriden Panel beantworten Teilnehmende Fragen zu gesellschaftlich relevanten Themen gemeinsam mit KI-Systemen. Wenn Sie teilnehmen, unterstützt Sie die KI bei der Beantwortung von Fragen,  ergänzt Informationen oder antwortet an Ihrer Stelle. Die KI verarbeitet dazu Informationen aus früheren Antworten und lernt aus Ihren Angaben.\\

\noindent\textit{[Dimension 1: Sensitivity of data, low vs. high]}

\noindent [Low] Die KI verarbeitet nur allgemeine Informationen, zum Beispiel zu Ihrem Alter, Geschlecht, Ihren Hobbys oder Interessen. 

\noindent [High] Die KI verarbeitet sensible persönliche Informationen, zum Beispiel zu Ihrer Gesundheit, Sexualität, Religion oder Ihren politischen Einstellungen. \\

\noindent\textit{[Dimension 2: AI autonomy \& Sensitivity of simulation, low vs. high]}

\noindent [Low] Die KI wird nur auf Ihre direkte Anweisung aktiv und handelt nicht selbstständig. Die KI macht Ihnen Vorschläge für mögliche Antworten, die Sie annehmen oder ablehnen können. Sie selbst entscheiden, welche Fragen von der KI beantwortet werden. 

\noindent [High] Die KI handelt selbstständig ohne Ihre direkte Anweisung. Die KI übernimmt die Beantwortung von Fragen für Sie. Die KI entscheidet, welche Fragen von ihr beantwortet werden.\\ 

\noindent\textit{[Dimension 3: Financial compensation, low vs. high]}

\noindent [Low] Für Ihre Teilnahme erhalten Sie 13,90 Euro pro Stunde. Das entspricht dem Mindestlohn. 

\noindent [High] Für Ihre Teilnahme erhalten Sie erhalten 20,85 Euro pro Stunde. Das ist 50\% mehr als der Mindestlohn.\\

\noindent Ihre Daten werden nach hohen Datenschutzstandards verarbeitet. Sie werden ausführlich darüber informiert, wie Ihre Daten gespeichert und genutzt werden. Sie können Ihre Teilnahme am hybriden Panel jederzeit beenden und die Löschung Ihrer Daten verlangen.

\noindent Wie wahrscheinlich wäre es, dass Sie an einem solchen hybriden Panel teilnehmen?

\paragraph{Response options:} sehr unwahrscheinlich - eher unwahrscheinlich - eher wahrscheinlich - sehr wahrscheinlich

\subsubsection{Attitudes towards artificial intelligence scale (ATTARI-12)}

\paragraph{Instruction:}
Im Folgenden interessieren wir uns für Ihre Einstellungen gegenüber Künstlicher Intelligenz (KI). Künstliche Intelligenz kann Aufgaben ausführen, die üblicherweise menschliche Intelligenz erfordern. Im äußersten Fall befähigt KI Maschinen dazu, selbstständig und ähnlich dem Menschen, ihre Umwelt wahrzunehmen, zu handeln, zu lernen und sich anzupassen. Künstliche Intelligenz kann Teil eines Computers oder einer Onlineplattform sein – man kann ihr aber auch in verschiedenen anderen technischen Geräten, wie etwa Robotern, begegnen. Bitte geben Sie an, inwieweit Sie den folgenden Aussagen zustimmen. Es gibt dabei keine richtigen oder falschen Antworten.

\begin{longtable}{|c|p{6cm}|c|c|}
\hline
 & \textbf{Formulierung} & \textbf{Facette} & \textbf{Valenz} \\
\hline
\endfirsthead

\hline
 & \textbf{Formulierung} & \textbf{Facette} & \textbf{Valenz} \\
\hline
\endhead

% Table content starts here
\textbf{1} & Künstliche Intelligenz wird die Welt verbessern. & Kognitiv & Positiv \\
\hline
\textbf{2} & Ich habe starke negative Emotionen gegenüber künstlicher Intelligenz. & Affektiv & Negativ (reverse-coded) \\
\hline
\textbf{3} & Ich möchte Technologien nutzen, die auf künstlicher Intelligenz basieren. & Behavioral & Positiv \\
\hline
\textbf{4} & Künstliche Intelligenz hat mehr Nachteile als Vorteile. & Kognitiv & Negativ (reverse-coded) \\
\hline
\textbf{5} & Ich freue mich auf zukünftige Entwicklungen im Bereich künstliche Intelligenz. & Affektiv & Positiv \\
\hline
\textbf{6} & Künstliche Intelligenz bietet Lösungen für viele globale Probleme. & Kognitiv & Positiv \\
\hline
\textbf{7} & Ich bevorzuge Technologien, die keine künstliche Intelligenz beinhalten. & Behavioral & Negativ (reverse-coded) \\
\hline
\textbf{8} & Ich fürchte mich vor künstlicher Intelligenz. & Affektiv & Negativ (reverse-coded) \\
\hline
\textbf{9} & Ich würde mich eher für eine Technologie mit künstlicher Intelligenz entscheiden als für eine ohne. & Behavioral & Positiv \\
\hline
\textbf{10} & Künstliche Intelligenz verursacht eher Probleme, anstatt sie zu lösen. & Kognitiv & Negativ (reverse-coded) \\
\hline
\textbf{11} & Wenn ich an künstliche Intelligenz denke, habe ich hauptsächlich positive Gefühle. & Affektiv & Positiv \\
\hline
\textbf{12} & Ich möchte mit Technologien, die auf künstlicher Intelligenz beruhen, lieber nichts zu tun haben. & Behavioral & Negativ (reverse-coded) \\
\hline
\end{longtable}

\paragraph{Response options:} stimme überhaupt nicht zu – stimme eher nicht zu – weder noch – stimme eher zu – stimme voll zu 

\subsubsection{Political leaning}

\paragraph{Instruction:} In der Politik reden die Leute häufig von ``links'' und ``rechts''. Wo würden Sie sich selbst einordnen?

\paragraph{Response options:}
\noindent
\textbf{links} \quad 1 \ 2 \ 3 \ 4 \ 5 \ 6 \ 7 \ 8 \ 9 \ 10 \ 11  \quad \textbf{rechts} - weiß nicht

\subsubsection{Gender}

\paragraph{Instruction:}
Welches Geschlecht haben Sie?

\paragraph{Response options:} Männlich - Weiblich - Divers

\subsubsection{Education}

\paragraph{Instruction:}
Welchen höchsten allgemeinbildenden Schulabschluss haben Sie?\\
Anmerkung: Wenn Sie Ihren höchsten Schulabschluss im Ausland gemacht haben, versuchen Sie sich bitte den vorgegebenen Kategorien zuzuordnen. 

\paragraph{Response options:}
Noch keinen – Schüler/in - Von der Schule abgegangen ohne Schulabschluss - Hauptschulabschluss (Volksschulabschluss) oder gleichwertiger Abschluss - Polytechnische Oberschule der DDR mit Abschluss der 8. oder 9. Klasse - Realschulabschluss (Mittlere Reife) oder gleichwertiger Abschluss - Polytechnische Oberschule der DDR mit Abschluss der 10. Klasse - Fachhochschulreife - Abitur/Allgemeine oder fachgebundene Hochschulreife (Gymnasium beziehungsweise EOS, auch EOS mit Lehre) - Einen anderen Schulabschluss, und zwar:

\end{document}